\documentclass{article}
\usepackage{spconf,amsmath,epsfig,amssymb}
\usepackage{graphicx}
\usepackage{amsmath}
\usepackage{amssymb}
\usepackage{booktabs}
\usepackage{multicol}
\usepackage{multirow}
\let\OLDthebibliography\thebibliography
\renewcommand\thebibliography[1]{
  \OLDthebibliography{#1}
  \setlength{\parskip}{0pt}
  \setlength{\itemsep}{0pt plus 0.3ex}
}

\begin{document}\sloppy

\def\x{{\mathbf x}}
\def\L{{\cal L}}


\title{Invertible Image Dataset Protection}
%
\name{Kejiang Chen$^{1\star}$, Xianhan Zeng$^{2\star}$, Qichao Ying$^{2}$, Sheng Li$^{2}$, Zhenxing Qian$^{2\ast}$\thanks{The first two authors contribute equally. $^{\ast}$Corresponding author.}\thanks{This work was supported in part by the National Natural Science Foundation of China under Grants U20B2051, 62072114, U20A20178.} and Xinpeng Zhang$^{2}$}
\address{$^{1}$University of Science and Technology of China, Anhui, China\\
$^{2}$Fudan University, Shanghai, China\\
chenkj@ustc.edu.cn, \{xhzeng20, qcying20, zxqian, zhangxinpeng\}@fudan.edu.cn\\
}

\maketitle

\begin{abstract}

Deep learning has achieved enormous success in various industrial applications. 
Companies do not want their valuable data to be stolen by malicious employees to train pirated models. Nor do they wish the data analyzed by the competitors after using them online. We propose a novel solution for dataset protection in this scenario by robustly and reversibly transform the images into adversarial images. We develop a reversible adversarial example generator (RAEG) that introduces slight changes to the images to fool traditional classification models. Even though malicious attacks train pirated models based on the defensed versions of the protected images, RAEG can significantly weaken the functionality of these models. Meanwhile, the reversibility of RAEG ensures the performance of authorized models. Extensive experiments demonstrate that RAEG can better protect the data with slight distortion against adversarial defense than previous methods. 

\end{abstract}

\section{Introduction}
Deep learning and neural network models have achieved great success in various fields including image classification and face recognition. 
Such superior performance causes concerns for both individuals and companies. On the one hand, individuals concern that their data will be analyzed by unauthorized models, leading to privacy leakage. On the other hand, large-scale datasets are necessary to train a satisfying deep model. However, data collecting and labelling are time-consuming and expensive. Therefore, datasets are always seemed as digital properties. Organizations, companies and researchers may charge for their carefully labeled or annotated private datasets and only those authorized are expected to use these datasets.
However, unauthorized malicious employees may steal the dataset and send it to the company's competitors, resulting in economic losses for the company. As a result, it is vital to protect the data from both instance-level as well dataset-level. 
 

Adversarial Examples (AE) ~\cite{szegedy2013intriguing,carlini2017towards} are specialised in confusing neural networks to make unreliable decisions. The generated examples are indistinguishable with normal ones to the human eye but function drastically different. However, since the generation of AE is gradually irreversible, it also bans the legal users from utilizing the resources. Reversible Adversarial Example (RAE) ~\cite{liu2018reversible} aims to alleviate this issue by reverting AE using reversible data hiding (RDH). The method first generates the adversarial image using existing AE methods, then embeds the adversarial perturbation into the adversarial image, and generates the stego image using RDH. Due to the characteristic of reversibility, the adversarial perturbation and the original image can be recovered. 
\begin{figure}[t]
	\centering
	\includegraphics[width=0.48\textwidth]{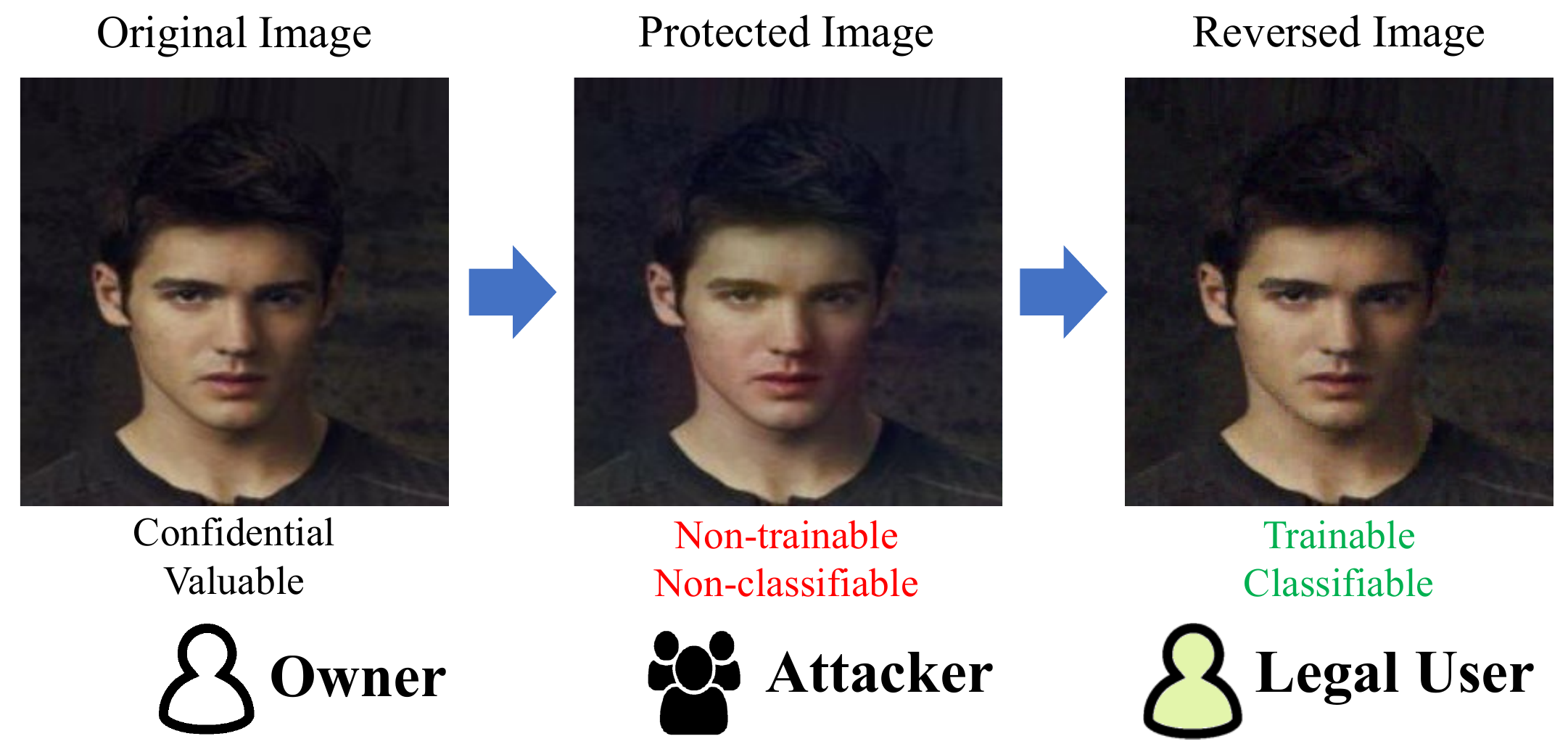}
	\caption{A practical example of RAEG. The data owner transforms his valuable images into non-recognizable, non-trainable adversarial images, despite using defensive methods like JPEG compression or ComDefend \cite{jia2019comdefend}. Legal users can revert the protection process with trivial loss.}
	\label{image_teaser}
\end{figure}
Yin \emph{et al.}~\cite{yin2019reversible} adopted Reversible Image Transform (RIT) instead of histogram shifting to improve the performance of RAE. 
However, RAE and RIT are fragile to adversarial defense methods, including Gaussian filtering, JPEG compression, etc. Therefore, the malicious attackers can easily obtain an approximate version of the original images using adversarial defense.

In this paper, we propose Reversible Adversarial Example Generation (RAEG) via an invertible neural network (INN), which takes the defense operations into consideration in the process of generation. In detail, the original image is fed into a U-shaped invertible neural network to obtain the adversarial image. Through the backward process of INN, we recover the original image with trivial loss. 
To enhance the robustness of RAEG against defensive operations, we mimic typical image processing operations and require that the protected images through these defenses still capable of fooling networks. 
Several classifiers with typical architecture are selected as targets for RAEG to attack with. Successful data protection model should enforce both the generated protected images and the attacked images to mislead the targeted models.
Figure~\ref{image_teaser} illustrates a practical application where the images are perturbed by RAEG for copyright protection. Even though the attacker processes the images using traditional blurring process or adversarial defense methods, he cannot successfully recover the images for network training. However, the authenticated receivers can successfully revert the protection process and train their networks based on the recovered images. Therefore, the data owner is not required to launch an unnecessary extra dataset transmission which is time and network consuming. 

Extensive experiments demonstrate that RAEG can better protect the data with slight distortion against image purifying operations than previous methods. The models trained upon protected images or their defensed versions are mal-functioned compared to normal ones. Besides, the proposed method can almost perfectly restore the original image, while the previous methods can only restore a blurry version.
Our main contributions are summarized as follows:
\begin{itemize}
    \item We propose a reversible adversarial example generation method based on invertible neural network to protect the dataset. To the best of our knowledge, this is the first work to introduce invertible neural network for generating adversarial examples.
    
    \item Considering defense operations, e.g., JPEG compression, we utilize the neural network to mimic these operations and force the generator to resist these operations.
    
   \item Extensive experiments show that the quality of adversarial images is considerable, and the ability of misleading the classification model with defense operations is better than traditional RAE methods.
\end{itemize}

\begin{figure*}[t]
	\centering
	\includegraphics[width=1.0\textwidth]{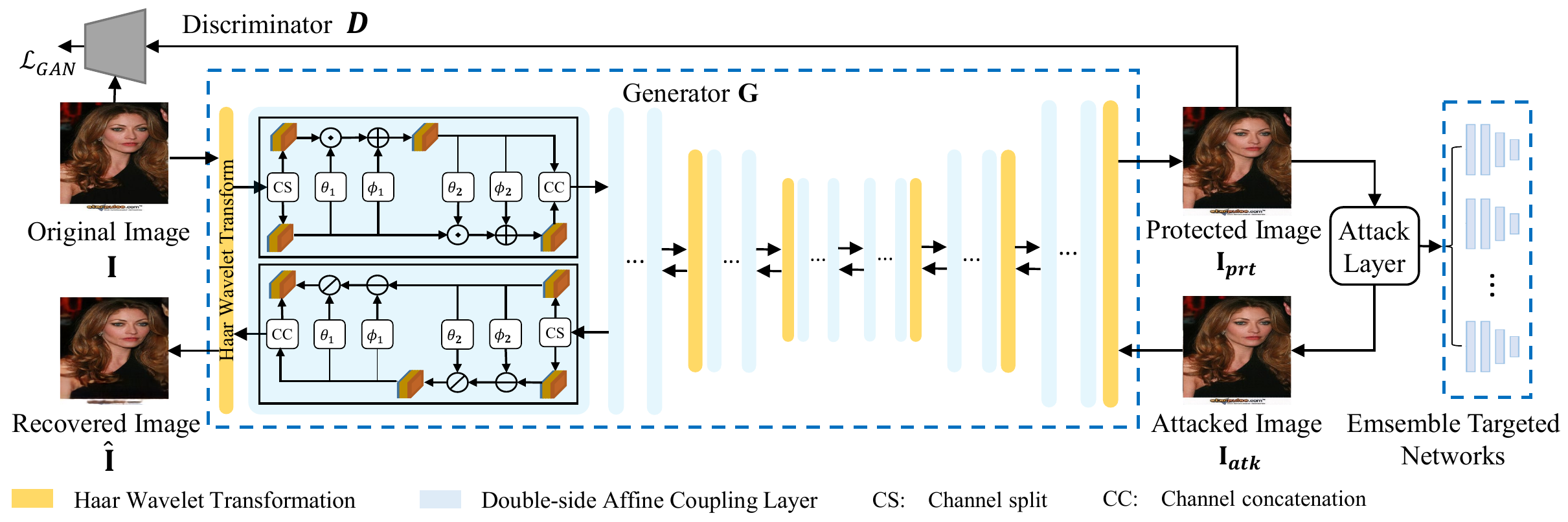}
	\caption{Network overview of RAEG. The generator transforms the original image into the protected image. The attack layer simulates the most commonly seen image processing attacks. We ensemble several classifiers as the joint targeted networks that the generated images should evade from. The inverse process of the generator recovers the original image.}
	\label{image_framework}
\end{figure*}

\section{Related work}

\subsection{Adversarial Attack and Defense}

The concept of the adversarial attack is first proposed by Szegedy et al.~\cite{szegedy2013intriguing}. 
IGSM~\cite{kurakin2016adversarial} iteratively uses the gradients of the loss with respect to the input image to create a new image that maximizes the loss.  C\&W~\cite{carlini2017towards} is an efficient optimization objective for iteratively finding the
adversarial examples with the smallest perturbation. Many other AE methods have been proposed in the past decades \cite{zhong2021undetectable}.

There exist different adversarial defense methods. Most representative methods are based on image denoising, e.g., ComDefend~\cite{jia2019comdefend} and image reconstrution, e.g., DIPDefend~\cite{dai2020dipdefend}.
Besides, traditional image processings such as JPEG compression, Gaussian filter can also serve as lightweight denoising methods. Adversarial training is another powerful technique for defense. To counter the defensive methods, there are also some robust AE methods \cite{shi2021generating}, but the perturbation is much larger and the robustness is not generalized.

\subsection{Reversible Adversarial Example}



Liu et al.~\cite{liu2018reversible} aim at to restoring the original image from the reversible adversarial image using RDH method. 
In detail, the adversarial image is generated by the existing AE methods, e.g. C\&W. Then, RAE try to hide the difference into the adversarial image using RDH. Due to the limited capacity of RDH, the information of the difference is usually too large, and they alternatively downsample the images and hide the resized difference. \textit{Therefore, the original image cannot be perfect restored. Besides, the adversarial defense methods can easily invalidate the RAE.}

To perfectly restore the original image, Yin et al.~\cite{yin2019reversible} introduce reversible image transform (RIT) for generating reversible adversarial example. The essence of RIT is permuting the blocks, e.g. $2\times2$, of the original image to match the adversarial image. Then some additional information, e.g., block index, should be embedded into the permuted image using RDH to generate the final image. It can be easily inferred that the permutation causes large distortion, e.g., severe block artifacts. \textit{The adversarial ability of RIT degrades a lot resulting from the distortion. }

\section{Method}


\subsection{Overview}
The sketch of our RAEG framework is presented in Figure \ref{image_framework}, which consists of a generator, an attack layer for defense simulation, several pre-trained targeted classifiers and a discriminator. The generator $\mathbf{G}$ is built on an INN, where given $\mathbf{I}$, we generate the protected version $\mathbf{I}_{\emph{prt}}$ using the forward pass. We introduce an attack layer to simulate the real-world image redistribution stage as well as enhance the robustness of the generated adversarial example.
Afterwards, we emsemble several pre-trained classifiers as the joint targeted networks. We require that both $\mathbf{I}_{\emph{atk}}$ and $\mathbf{I}_{\emph{prt}}$ should mislead the joint prediction of the emsemble models. 
In the backward pass, we inversely run $\mathbf{G}$ to recover $\mathbf{I}$.
Finally, we introduce a discriminator to keep high visual quality of the adversarial image.



\subsection{Network Architecture}
\noindent\textbf{The Generator. }
As shown in Figure~\ref{image_framework}, the generator $\mathbf{G}$ is composed of three stacked downscaling
modules and three upsampling modules. Each module contains a Haar downsampling or upsampling transformation layer and four double-sided affine coupling layers.
After the Haar wavelet transformation, the input image or the feature map is decomposed into one low-frequency representation and three high-frequency representations in the vertical, horizontal, and diagonal direction. The wavelet transform is an invertible symmetric transformation, which will not affect the following invertible operations.

Afterwards, the frequency representations are fed to
double-side affine coupling (DSAC) block, which
are the crucial parts in the invertible process. 
DSAC block first splits the input image or the feature map $\boldsymbol{x}^i$ into two parts, denoted as $\boldsymbol{x}_1^i$ and $\boldsymbol{x}_2^i$.
Then, we apply double-side affine transformations to both $\boldsymbol{x}_1^i$ and $\boldsymbol{x}_2^i$:
\begin{equation}
    \begin{split}
        \boldsymbol{x}_1^{i+1} &= \boldsymbol{x}_1^i \odot \exp(\theta_1(\boldsymbol{x}_2^i)) + \phi_1(\boldsymbol{x}_2^i) \\
        \boldsymbol{x}_2^{i+1} &= \boldsymbol{x}_2^i \odot \exp(\theta_2(\boldsymbol{x}_1^{i+1})) + \phi_2(\boldsymbol{x}_1^{i+1}),
    \end{split}
\end{equation}
where $\theta(\cdot)$ and $\phi(\cdot)$ produce the scale and shift coefficients. The output sub-band feature maps of each block are then concatenated into a complete output feature map, denoted as $(\boldsymbol{x}_{1}^{i+1}, \boldsymbol{x}_{2}^{i+1})$.
In the reverse process, the input feature map is first split into two sub-band feature maps $\boldsymbol{x}_{1}^{i+1}$ and $\boldsymbol{x}_{2}^{i+1}$, and then use double-side affine transformations to get two reversed output sub-band feature maps $\boldsymbol{x}_1^i$ and $\boldsymbol{x}_2^i$:
\begin{equation}
    \begin{split}
        \boldsymbol{x}_2^i &= (\boldsymbol{x}_2^{i+1} - \phi_2(\boldsymbol{x}_1^{i+1})) \oslash \exp(\theta_2(\boldsymbol{x}_1^{i+1})) \\
        \boldsymbol{x}_1^i &= (\boldsymbol{x}_1^{i+1} - \phi_1(\boldsymbol{x}_2^i)) \oslash \exp(\theta_1(\boldsymbol{x}_2^i)),
    \end{split}
\end{equation}
where $\oslash$ denotes element-wise division. Finally, we concatenate two output sub-band feature maps into the reversed feature map $\boldsymbol{x}^i$. Noted that $\theta(\cdot)$ and $\phi(\cdot)$ are not required to be invertible. We use the residual block to construct the sub-nets. Spectral Normalization (SN)~\cite{miyato2018spectral} instead of the traditional batch normalization is adopted in each block, since it helps stabilizing the training.



\noindent\textbf{The Defense Simulation.}
Similar to~\cite{zhu2018hidden}, we mimic typical image preprocessing by differentiable layers. Our defense layer contains five types of typical digital attacks, including Gaussian noise, Gaussian blurring, scaling, random cropping and JPEG compressing. 
Since JPEG simulation in previous works are reported to have over-fitting problem in that the compression mode is always fixed, we propose to generated the pseudo-JPEG images $\mathbf{I}_{\emph{jpg}}$ by interpolating the results among different methods with varied quality factors.
\begin{equation}
\mathbf{I}_{\emph{jpg}}=\sum_{\mathcal{J}_{k}\in \mathcal{J}}\sum_{QF_{l}\in[10,100]}\epsilon\cdot\mathcal{J}_{k}(\mathbf{I}_{prt}, QF_{l}),
\end{equation}
where $QF$ stands for the quality factor, $\sum_{\mathcal{J}_{k}\in \mathcal{J}}\epsilon=1$ and $\mathcal{J} \in $\{JPEG-SS\cite{liu2021jpeg}, JPEG-Mask\cite{zhu2018hidden}, MBRS\cite{jia2021mbrs}\}.

\noindent\textbf{The Ensemble Targeted Networks. }
We employ four targeted classifiers with different architectures, i.e., VGG16, ResNet-50, ResNet-101 and DenseNet-121. The networks are pre-trained on clean $\mathbf{I}$.

\noindent\textbf{The Discriminator. }
Finally, we introduce a discriminator $\mathbf{D}$ to improve the quality of $\mathbf{I}_{\emph{prt}}$, which distinguishes the generated images $\mathbf{I}_{\emph{prt}}$ from the original image $\mathbf{I}$.

\begin{figure}[!t]
	\centering
	\includegraphics[width=0.49\textwidth]{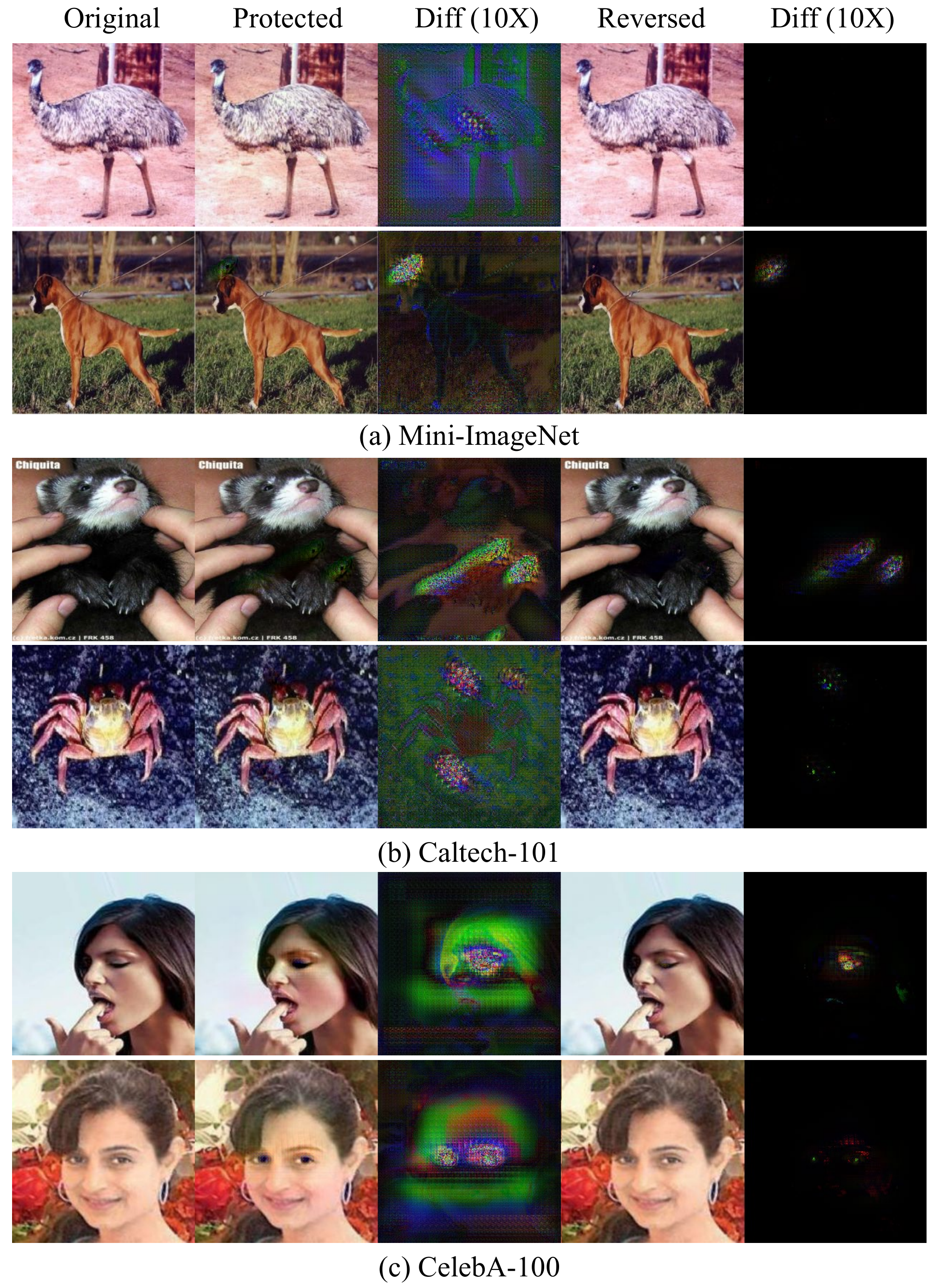}
	\caption{Inversion test of RAEG. We perform different kinds of attacks on the protected images before reconstruction, e.g., JPEG with QF=70 on (a), Gaussian Blur on (b) and rescaling on (c). RAEG can still reverse the images despite the presence of the attacks.}
	\label{image_comparison}
\end{figure}
\begin{figure}[!t]
	\centering
	\includegraphics[width=0.49\textwidth]{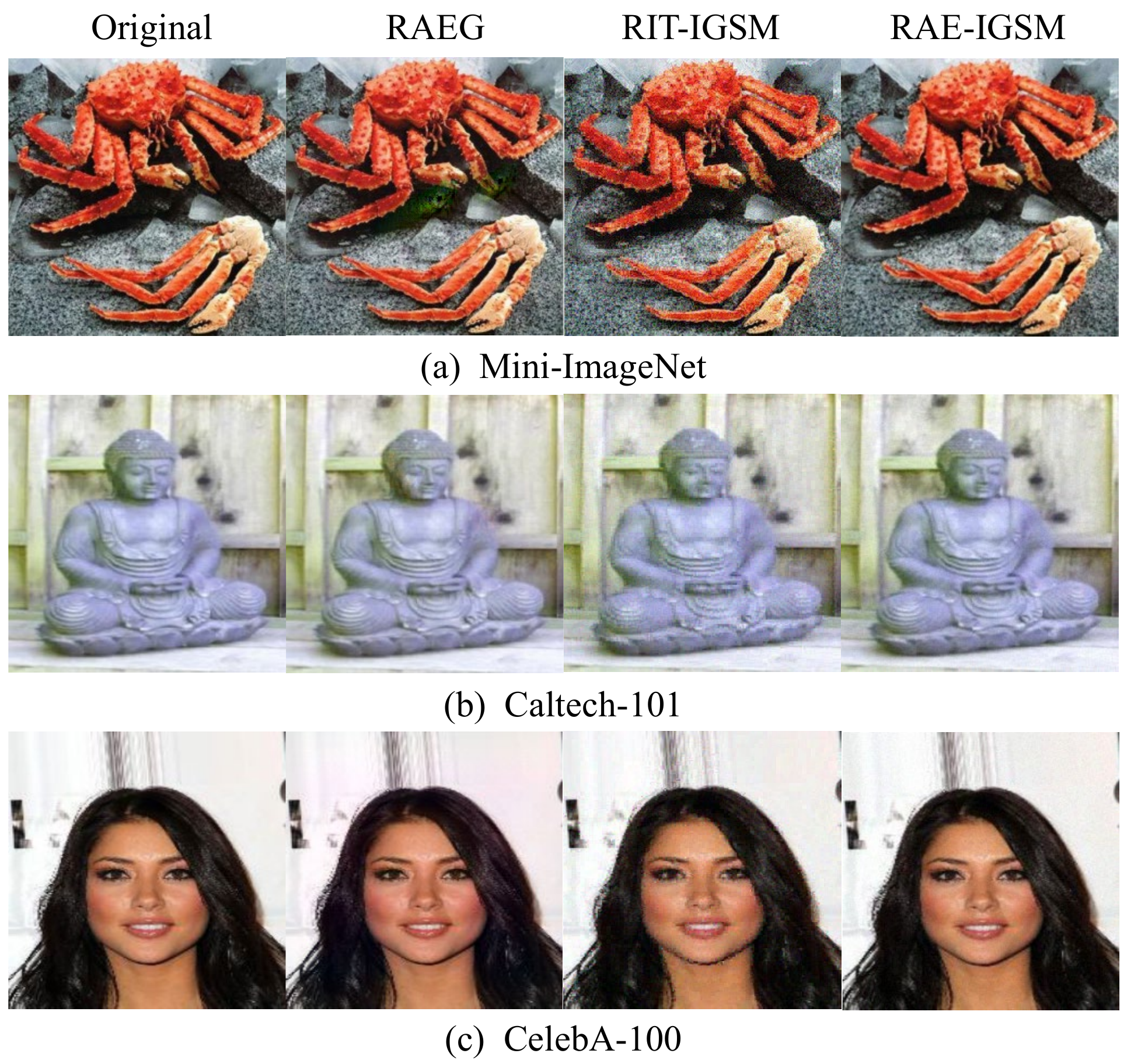}
	\caption{Showcase of protected images among RAE-IGSM, RIT-IGSM and RAEG. The visual quality of RAE-generated images are of good quality yet the robustness is much lower (specified in Table~\ref{table_comparison}).}
	\label{image_protected}
\end{figure}

\subsection{Loss Functions}
The objective functions include the reconstruction loss $\mathcal{L}_{prt}$, the classification loss $\mathcal{L}_{\emph{loc}}$ and the adversarial loss $\mathcal{L}_{\emph{adv}}$.
In the following, $\alpha$, $\beta$, $\gamma$, $\delta$ and $\epsilon$ are hyper-parameters.

\noindent\textbf{Reconstruction Loss.}
We employ a protection loss $\mathcal{L}_{prt}$ and a recovery loss $\mathcal{L}_{\emph{rec}}$ respectively to encourage $\mathbf{I}_{\emph{prt}}$ and the recovered image $\hat{\mathbf{I}}$ to resemble $\mathbf{I}$. 
For $\mathcal{L}_{\emph{rec}}$, we have \begin{equation}
    \mathcal{L}_{prt}=\lVert\mathbf{I}-\mathbf{I}_{\emph{prt}}\rVert_{1}.
\end{equation}
For $\mathcal{L}_{\emph{prt}}$, we additionally employ the perceptual loss $\mathcal{L}_{\emph{per}}$ by employing a VGG16 feature extractor. We compare the feature maps of $\mathbf{I}$ and $\mathbf{I}_{prt}$ from the second max-pooling layer. 
\begin{equation}
    \mathcal{L}_{per} = \alpha\cdot\sum_{i=1}^N \frac{1}{H \cdot W \cdot C}|\phi_{pool_i}^{gt} - \phi_{pool_i}^{pred}|_1+\lVert\mathbf{I}-\hat{\mathbf{I}}\rVert_{1},
\end{equation}
where $H, W, C$ refer to the height, weight and channel number of features.

\noindent\textbf{Discriminator Loss.}
For the discriminator loss, we update $\mathcal{D}$ by minimizing the binary classification loss:
\begin{equation}
    \mathcal{L}_{dis} = \mathbb{E}_{\mathbf{I}} \log (\mathcal{D}(\mathbf{I})) + \mathbb{E}_{\mathbf{I}_{\emph{prt}}} \log (1 - \mathbf{D}(\mathbf{I}_{\emph{prt}})).
\end{equation}
The generator aims to fool the discriminator to make wrong predictions on whether an image is an original image or a generated image. The GAN loss $\mathcal{L}_\emph{GAN}$ is therefore:
\begin{equation}
\mathcal{L}_\emph{GAN}=\mathbb{E}_{\mathbf{I}_{\emph{prt}}} \log ( \mathcal{D}(\mathbf{I}_{\emph{prt}})).
\end{equation}


\noindent\textbf{Classification Loss.}
We use the negative cross entropy (CE) loss to encourage $\mathbf{I}_{\emph{prt}}$ to be wrongly predicted by the targeted models. Considering that the $\ell_1$ distance term alone might not be powerful enough to recover the trivial details of $\mathbf{I}$, we add an additional CE loss to further encourage $\hat{\mathbf{I}}$ to be correctly classified. Therefore, the classification loss is defined as:
\begin{equation}
    \mathcal{L}_\emph{cls}=\emph{CE}(\emph{f}(\hat{\mathbf{I}}),l)-\epsilon\cdot~\emph{CE}(\emph{f}(\mathbf{I}_\emph{prt}),l).
\end{equation}

\noindent\textbf{Total Loss.}
The total loss is defined as follow:
\begin{equation}
    \mathcal{L} = \mathcal{L}_{prt} + \beta\cdot \mathcal{L}_{rev} + \gamma\cdot\mathcal{L}_{cls} + \delta \mathcal{L}_{\emph{GAN}}.
\end{equation}

\begin{table*}
\footnotesize
\centering
\caption{Comparison of accuracy on pretrained ResNet-50/DenseNet-121 using RAE\cite{liu2018reversible}, RIT\cite{yin2019reversible} and RAEG.}
\begin{tabular}{c|c|c|cc|c|c|c|c|c} 
\hline
\multirow{2}{*}{Dataset}      & \multirow{2}{*}{Method} & No     & \multicolumn{2}{c|}{JPEG} & \multirow{2}{*}{Resize}  & Gaussian & Medium  & \multirow{2}{*}{ComDefend}     & \multirow{2}{*}{DIPDefend}   \\
                              &                          & Attack & QF=90 & QF=50     &   & Noise    & Filter   & &                    \\ 
\hline
\multirow{4}{*}{\begin{tabular}[c]{@{}c@{}}Caltech\\-101\end{tabular}}             & RAEG         & 0.479 / 0.495     & 0.587 / 0.570   & 0.752 / \textbf{0.721}   & \textbf{0.615 / 0.701}     & \textbf{0.607 / 0.732}  & \textbf{0.504 / 0.672}     & \textbf{0.632 / 0.709}   & \textbf{0.587 / 0.704}                 \\
& RAE-IGSM     & \textbf{0.076 / 0.063}     & \textbf{0.376 / 0.394}   & 0.890 / 0.873   & 0.755 / 0.712     & 0.816 / 0.822  & 0.823 / 0.799     & 0.748 / 0.752   & 0.843 / 0.835                 \\
& RAE-CW       & 0.656 / 0.673     & 0.955 / 0.958   & 0.961 / 0.967   & 0.950 / 0.961     & 0.957 / 0.942  & 0.880 / 0.868     & 0.937 / 0.921   & 0.891 / 0.903                 \\
& RIT-IGSM     & 0.679 / 0.850     & 0.697 / 0.854   & \textbf{0.693} / 0.854   & 0.677 / 0.872     & 0.729 / 0.888  & 0.693 / 0.829     & 0.779 / 0.891   & 0.747 / 0.893                 \\
\hline
\multirow{4}{*}{\begin{tabular}[c]{@{}c@{}}CelebA\\-100\end{tabular}}              & RAEG         & 0.060 / 0.093     & 0.070 / 0.090   & \textbf{0.073 / 0.083}   & 0.103 / 0.127     & \textbf{0.093 / 0.113}  &\textbf{0.070 / 0.083}     & \textbf{0.080 / 0.093}   & \textbf{0.080 / 0.089}                 \\
& RAE-IGSM     & \textbf{0.003 / 0.001}     & \textbf{0.003 / 0.003}   & 0.275 / 0.267   & \textbf{0.053 / 0.066}     & 0.113 / 0.125  & 0.343 / 0.359     & 0.257 / 0.298   & 0.387 / 0.365                 \\
& RAE-CW       & 0.493 / 0.452     & 0.762 / 0.733   & 0.796 / 0.762   & 0.773 / 0.745     & 0.752 / 0.736  & 0.657 / 0.662     & 0.710 / 0.694   & 0.656 / 0.677                 \\
& RIT-IGSM     & 0.114 / 0.442     & 0.380 / 0.428   & 0.137 / 0.421   & 0.127 / 0.391     & 0.174 / 0.515  & 0.204 / 0.522     & 0.224 / 0.539   & 0.197 / 0.572                 \\
\hline
\multirow{4}{*}{\begin{tabular}[c]{@{}c@{}}Mini-\\ImageNet\end{tabular}}           & RAEG         & \textbf{0.068 / 0.092}     & \textbf{0.082 / 0.113}   & \textbf{0.171 / 0.183}   & \textbf{0.117 / 0.148}     & \textbf{0.129 / 0.118}  & \textbf{0.087 / 0.124}     & \textbf{0.099 / 0.124}   & \textbf{0.101 / 0.183}                 \\
& RAE-IGSM     & 0.192 / 0.225     & 0.488 / 0.473   & 0.819 / 0.835   & 0.687 / 0.691     & 0.770 / 0.796  & 0.790 / 0.788     & 0.796 / 0.802   & 0.823 / 0.819                 \\
& RAE-CW       & 0.434 / 0.451     & 0.929 / 0.917   & 0.923 / 0.940   & 0.902 / 0.925     & 0.924 / 0.912  & 0.859 / 0.863     & 0.900 / 0.878   & 0.870 / 0.856                 \\
& RIT-IGSM     & 0.572 / 0.805     & 0.586 / 0.791   & 0.623 / 0.791   & 0.609 / 0.813     & 0.751 / 0.879  & 0.655 / 0.799     & 0.755 / 0.860   & 0.751 / 0.880                 \\

\hline
\end{tabular}
\label{table_comparison}
\end{table*}

\section{Experiment}

\subsection{Experimental Settings}
We train the target classification networks with batch size 32 in advance and then freeze the parameters. We train the invertible framework for 50 epochs. The hyper-parameters are set as $\alpha=0.01, \beta=1, \gamma=0.005, \epsilon=2$ and $\delta=0.01$. Adam is adopted as the optimizer, and the batch size is 8. The learning rate is $10^{-4}$. The models are all trained on NVIDIA RTX 3090 and the training finishes in a week. 

\noindent\textbf{Dataset and Evaluation.} 
We evaluate our proposed RAEG on three typical datasets, namely, CelebA-100, Caltech-101 and Mini-ImageNet. 
We divide the images in each category or class into training set and test set at a ratio of $9:1$ and resize all the images to $256 \times 256$ pixels.
We employ peak signal to noise ratio (PSNR) and structural similarity (SSIM) in our quantitative image comparisons. We report the Top-1 accuracy during evaluation.

\noindent\textbf{Benchmark.}
There is no previous work that can simultaneously ensure the robustness of the adversary and provide invertibility. RAE and RIT embed the difference between the original and the adversarial image. However, most robust AE methods \cite{shi2021generating,zhong2021undetectable} introduces too much distortion, which exceeds the maximum capacity of state-of-the-art RDH schemes according to our experiments. Therefore, it is hard to simultaneously ensure robustness and invertibility based on RAE and RIT. We implement respectively RIT and RAE using IGSM and CW as AE and RDH-OVT \cite{zhang2012reversible} as RDH. 





\subsection{Qualitative and Quantitative Analysis}

\noindent\textbf{Qualitative Results.} 
Figure~\ref{image_comparison} and Figure~\ref{image_protected} provides examples of RAEG inversion test and RAEG-generated protected images where $\mathbf{I}$ are acceptably modified for protection. The perturbation is generally represented by contrast change or local area modification. With $\mathbf{I}$ kept secret, the overall quality of $\mathbf{I}_{\emph{prt}}$ is decent enough for normal usage of these images. The average PSNR between $\mathbf{I}$ and $\mathbf{I}_{\emph{prt}}$ in these examples is 28.12dB. In contrast, RIT cannot generate visual-satisfactory protected images in that there are noticable distortions. Please zoom in for details. The quality of generated image by RAE is the best, but the robustness of adversary is limited according to the following tests. In Figure~\ref{image_comparison}, the protected images cannot be categorized by traditional classifiers while the reversed images can, and $\hat{\mathbf{I}}$ is nearly same as the $\mathbf{I}$.

\noindent\textbf{Quantitative Comparisons.} 
Table~\ref{table-psnr-acc} presents the numerical results of our RAEG under various parameters. Their exist a trade-off between the adversary and the visual quality of the protected image. PSNR$\approx$28dB in RAEG  provides a considerable
trade-off, so we adopt this experimental setting during comparison. As for the recovered image $\hat{\mathbf{I}}$, the PSNR is around 40dB and SSIM is greater than 0.97.

Table~\ref{table_comparison} compares the accuracy of the target models on images generated by different RAE methods and RAEG. our RAEG signiﬁcantly outperforms RAE methods against typical attacks, including low quality JPEG compression, resize. Besides, we also report the anti-defense capability against two state-of-the-art methods, namely, ComDenfend \cite{jia2019comdefend} and DIPDefend \cite{dai2020dipdefend}. The results are also promising that the attackers cannot effectively increase the classification accuracy.

\noindent\textbf{Prevention of Training Pirated Models.} 
We further explore the performance of RAEG on protecting the whole dataset. We assume that the attacker can obtain the whole dataset from collecting enough $\mathbf{I}_{\emph{prt}}$, and he trains his own ResNet-50/ResNet-101/DenseNet-121 network based on these images. As can be seen from Table~\ref{table-retrain}, the three pirated models cannot be well trained on the protected dataset, even if he pre-processes $\mathbf{I}_{\emph{prt}}$ using ComDefend. In contrast, a verified recipient can train his networks well by inverting $\mathbf{I}_{\emph{prt}}$. The results clearly point out the effectiveness of RAEG to protect the whole dataset.   

\begin{table}[!t]
\footnotesize
\centering
\caption{Results of RAEG under various settings. $A$, $P$, $S$ respectively denote Accuracy, PSNR and SSIM.}
\begin{tabular}{c|c|c|c|c|c|c|c} 
\hline
Dataset & $A_{ori}$    & $A_{prt}$ & $A_{rev}$    & $P_{prt}$    & $P_{rev}$     & $S_{prt}$    & $S_{rev}$      \\
                             
\hline
\multirow{3}{*}{\begin{tabular}[c]{@{}c@{}}Caltech\\-101\end{tabular}}              & 0.953 & 0.654 & 0.944 & 30.504  & 45.499   & 0.900   & 0.989     \\
                                          & 0.980 & 0.479 & 0.977 & 27.256  & 45.260   & 0.848   & 0.985     \\
                                          & 0.956 & 0.166 & 0.941 & 24.167  & 41.684   & 0.728   & 0.963     \\ 
\hline
\multirow{3}{*}{\begin{tabular}[c]{@{}c@{}}CelebA\\-100\end{tabular}}             & 0.823 & 0.147 & 0.750 & 31.878  & 43.542   & 0.955   & 0.996     \\
                                          & 0.820 & 0.060 & 0.823 & 28.720  & 48.042   & 0.901   & 0.998     \\
                                          & 0.833 & 0.020 & 0.643 & 25.928  & 38.575   & 0.872   & 0.989     \\ 
\hline
\multirow{3}{*}{\begin{tabular}[c]{@{}c@{}}Mini-\\ImageNet\end{tabular}}            & 0.926 & 0.211 & 0.889 & 30.415  & 42.262   & 0.893   & 0.983     \\
                                  & 0.948 & 0.068 & 0.941 & 27.366  & 39.361   & 0.871   & 0.979     \\
                                  & 0.942 & 0.086 & 0.933 & 25.947  & 40.209   & 0.830   & 0.979     \\
\hline
\end{tabular}
\label{table-psnr-acc}
\end{table}

\begin{table}
\footnotesize
    \centering
    \caption{Training networks based on different datasets. The accuracies are from ResNet-50/ResNet-101/DenseNet-121.}
    \begin{tabular}{c|c|c}
        \hline
        \multirow{2}{*}{Dataset}  & \multicolumn{1}{c|}{Trained on $\mathbf{I}_{atk}$} & \multirow{2}{*}{Trained on $\hat{\mathbf{I}}$}  \\
       & using ComDefend &  \\
        \hline
        Caltech-101 & 0.473 / 0.552 / 0.591 & 0.974 / 0.971 / 0.967    \\
        \hline
        Mini-ImageNet & 0.526 / 0.565 / 0.597 & 0.935 / 0.955 / 0.952     \\
        \hline
        CelebA-100  &  0.243 / 0.252 / 0.260 & 0.805 / 0.807 / 0.801     \\
        \hline
    \end{tabular}
    \label{table-retrain}
\end{table}
\subsection{Ablation Study}
\begin{figure}[t]
	\centering
	\includegraphics[width=0.5\textwidth]{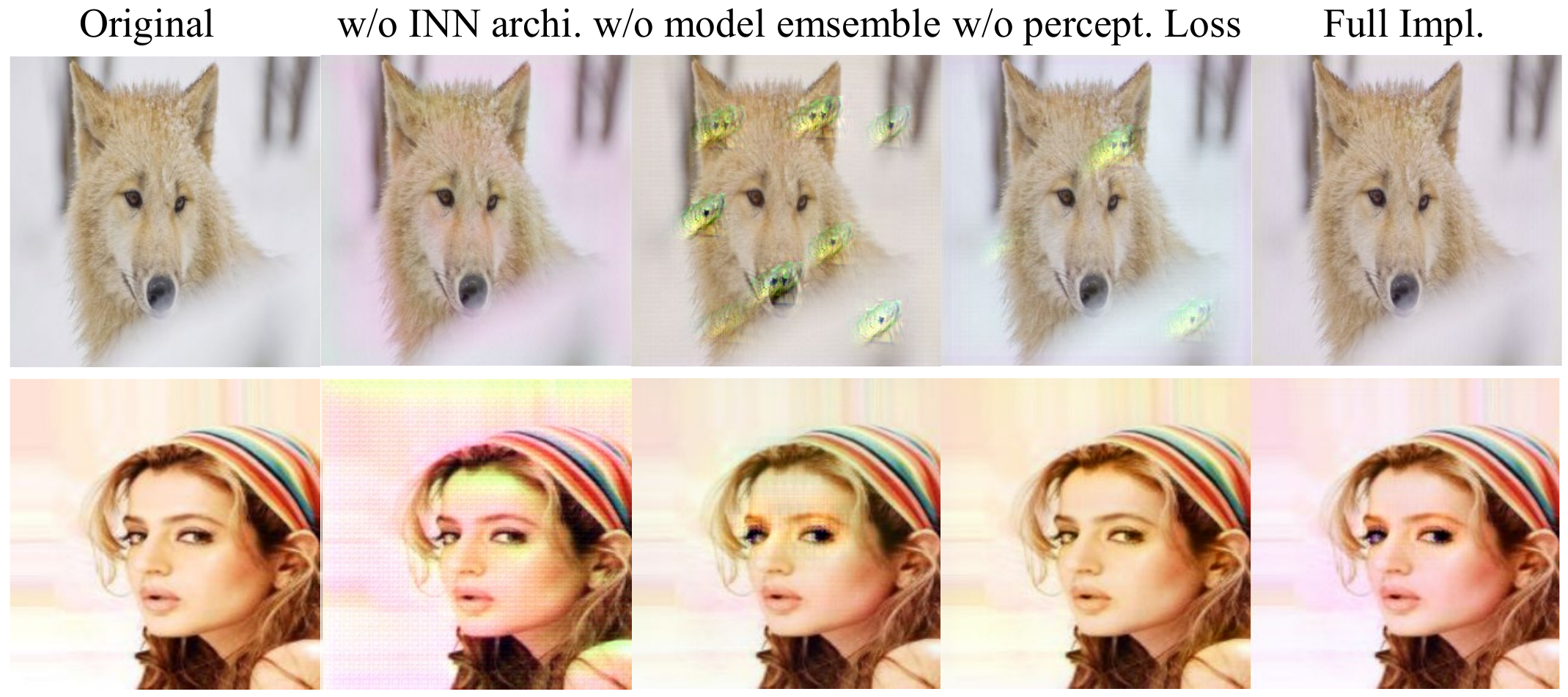}
	\caption{Ablation study on the quality of $\mathbf{I}_{\emph{prt}}$. The performances from the partial setups have noticeable defects.}
	\label{image_ablation}
\end{figure}
\begin{table}[t]
\footnotesize
    \centering
    \caption{Ablation study on robustness, visual quality and generalization of RAEG.}
    \begin{tabular}{c|c|c|c|c}
    \hline
    		Setting & $P_{rev}$ & $A_{prt}$ & $A_{tran}$ & $A_{def}$\\
    		\hline
    		$\mathbf{G}$ using ``Enc-Dec" architecture  & 31.37 & 0.479 & 0.569 & 0.803 \\
    		\hline
    		w/o discriminator & 37.53 & 0.241  & 0.265 & 0.491 \\
    		\hline
    		w/o VGG loss & 38.62 & 0.197  & 0.247 & 0.343\\
    			\hline
    		Training with one victim & $\textbf{42.61}$ & $0.124$ & $0.267$  & $0.673$ \\
    		\hline
    		Full Implementation & $40.14$ & $\textbf{0.093}$ & $\textbf{0.082}$  & $\textbf{0.122}$ \\
    		\hline
    	\end{tabular}
    	\label{table_ablation}
\end{table}

For space limit, we conduct ablation studies on Mini-ImageNet. The accuracies are reported based on a pretrained ResNet-50. Figure~\ref{image_ablation} and Table~\ref{table_ablation} respectively showcase and provide average results from several partial setups. For fair comparison, we let $P_{prt}\approx27$dB in each test.

\noindent\textbf{Effectiveness of INN Architecture.}
A typical alternative of INN is to model image protection and image recovery independently using Encoder-Decoder-based networks. However, invertible neural network learns stable invertible distribution mapping, where the forward and back propagation operations are in the same network. From Table~\ref{table_ablation}, we observe that INN-based RAEG provides a better performance.

\noindent\textbf{Influence of the Perceptual Loss and the Discriminator.}
As show in Table~\ref{table_ablation}, without the perceptual Loss and the discriminator, the overall performance will decline, and more visually unpleasant distortions appear. The results indicate that they do improve the visual quality.

\noindent\textbf{Investigation on the Target Model.} Training with one target model can improve the visual quality of $x_{rev}$, but it greatly degrades the robustness in the test stage, which highlights the necessity of training with emsemble targeted networks. The reason is that the attack model can easily find the weakness of a fixed classifier through iteration and utimately produces a fixed pattern. It results in poor performance generalization.



\section{Conclusion}
In this paper, we design a U-shaped invertible image transfer network to nullify the dataset by generating acceptable perturbations to alter their intrinsic properties. The reversibility of the image transfer network ensures the performance of authorized users. Extensive experiments demonstrate that the proposed method can effectively nullify the images with acceptable distortion. 

\bibliographystyle{IEEEbib}
\bibliography{icme2022template}

\begin{thebibliography}{10}

\bibitem{goodfellow2014explaining}
Ian~J Goodfellow, Jonathon Shlens, and Christian Szegedy,
\newblock ``Explaining and harnessing adversarial examples,''
\newblock {\em arXiv preprint arXiv:1412.6572}, 2014.

\bibitem{zhang2012reversible}
Xinpeng Zhang,
\newblock ``Reversible data hiding with optimal value transfer,''
\newblock {\em IEEE Transactions on Multimedia}, vol. 15, no. 2, pp. 316--325,
  2012.

\bibitem{zhong2021undetectable}
Nan Zhong, Zhenxing Qian, and Xinpeng Zhang,
\newblock ``Undetectable adversarial examples based on microscopical
  regularization,''
\newblock in {\em 2021 IEEE International Conference on Multimedia and Expo
  (ICME)}. IEEE, 2021, pp. 1--6.

\bibitem{dai2020dipdefend}
Tao Dai, Yan Feng, Dongxian Wu, Bin Chen, Jian Lu, Yong Jiang, and Shu-Tao Xia,
\newblock ``Dipdefend: Deep image prior driven defense against adversarial
  examples,''
\newblock in {\em ACM International Conference on Multimedia}, 2020, pp.
  1404--1412.

\bibitem{kim2018skewed}
Suah Kim, Xiaochao Qu, Vasily Sachnev, and Hyoung~Joong Kim,
\newblock ``Skewed histogram shifting for reversible data hiding using a pair
  of extreme predictions,''
\newblock {\em IEEE Transactions on Circuits and Systems for Video Technology},
  vol. 29, no. 11, pp. 3236--3246, 2018.

\bibitem{shin2017jpeg}
Richard Shin and Dawn Song,
\newblock ``{JPEG}-resistant adversarial images,''
\newblock in {\em NIPS 2017 Workshop on Machine Learning and Computer
  Security}, 2017, vol.~1.

\bibitem{shi2021generating}
Mengte Shi, Sheng Li, Zhaoxia Yin, Xinpeng Zhang, and Zhenxing Qian,
\newblock ``On generating {JPEG} adversarial images,''
\newblock in {\em IEEE International Conference on Multimedia and Expo}, 2021,
  pp. 1--6.

\bibitem{liu2021jpeg}
Kunlin Liu, Dongdong Chen, Jing Liao, Weiming Zhang, Hang Zhou, Jie Zhang,
  Wenbo Zhou, and Nenghai Yu,
\newblock ``{JPEG} robust invertible grayscale,''
\newblock {\em IEEE Transactions on Visualization and Computer Graphics}, 2021.

\bibitem{isola2017image}
Phillip Isola, Jun-Yan Zhu, Tinghui Zhou, and Alexei~A Efros,
\newblock ``Image-to-image translation with conditional adversarial networks,''
\newblock in {\em Proceedings of the IEEE conference on computer vision and
  pattern recognition}, 2017, pp. 1125--1134.

\bibitem{jia2019comdefend}
Xiaojun Jia, Xingxing Wei, Xiaochun Cao, and Hassan Foroosh,
\newblock ``Comdefend: An efficient image compression model to defend
  adversarial examples,''
\newblock in {\em Proceedings of the IEEE/CVF Conference on Computer Vision and
  Pattern Recognition}, 2019, pp. 6084--6092.

\bibitem{mao2017least}
Xudong Mao, Qing Li, Haoran Xie, Raymond~YK Lau, Zhen Wang, and Stephen
  Paul~Smolley,
\newblock ``Least squares generative adversarial networks,''
\newblock in {\em Proceedings of the IEEE international conference on computer
  vision}, 2017, pp. 2794--2802.

\bibitem{miyato2018spectral}
Takeru Miyato, Toshiki Kataoka, Masanori Koyama, and Yuichi Yoshida,
\newblock ``Spectral normalization for generative adversarial networks,''
\newblock {\em arXiv preprint arXiv:1802.05957}, 2018.

\bibitem{jia2021mbrs}
Zhaoyang Jia, Han Fang, and Weiming Zhang,
\newblock ``{MBRS}: Enhancing robustness of {DNN}-based watermarking by
  mini-batch of real and simulated {JPEG} compression,''
\newblock in {\em ACM International Conference on Multimedia}, 2021.

\bibitem{ioffe2015batch}
Sergey Ioffe and Christian Szegedy,
\newblock ``Batch normalization: Accelerating deep network training by reducing
  internal covariate shift,''
\newblock in {\em International conference on machine learning}. PMLR, 2015,
  pp. 448--456.

\bibitem{kurakin2016adversarial}
Alexey Kurakin, Ian Goodfellow, Samy Bengio, et~al.,
\newblock ``Adversarial examples in the physical world,'' 2016.

\bibitem{carlini2017towards}
Nicholas Carlini and David Wagner,
\newblock ``Towards evaluating the robustness of neural networks,''
\newblock in {\em IEEE Symposium on Security and Privacy}, 2017, pp. 39--57.

\bibitem{xiao2018spatially}
Chaowei Xiao, Jun-Yan Zhu, Bo~Li, Warren He, Mingyan Liu, and Dawn Song,
\newblock ``Spatially transformed adversarial examples,''
\newblock {\em arXiv preprint arXiv:1801.02612}, 2018.

\bibitem{liu2016delving}
Yanpei Liu, Xinyun Chen, Chang Liu, and Dawn Song,
\newblock ``Delving into transferable adversarial examples and black-box
  attacks,''
\newblock {\em arXiv preprint arXiv:1611.02770}, 2016.

\bibitem{su2019one}
Jiawei Su, Danilo~Vasconcellos Vargas, and Kouichi Sakurai,
\newblock ``One pixel attack for fooling deep neural networks,''
\newblock {\em IEEE Transactions on Evolutionary Computation}, vol. 23, no. 5,
  pp. 828--841, 2019.

\bibitem{moosavi2016deepfool}
Seyed-Mohsen Moosavi-Dezfooli, Alhussein Fawzi, and Pascal Frossard,
\newblock ``Deepfool: a simple and accurate method to fool deep neural
  networks,''
\newblock in {\em IEEE Conference on Computer Vision and Pattern Recognition},
  2016, pp. 2574--2582.

\bibitem{moosavi2017universal}
Seyed-Mohsen Moosavi-Dezfooli, Alhussein Fawzi, Omar Fawzi, and Pascal
  Frossard,
\newblock ``Universal adversarial perturbations,''
\newblock in {\em Proceedings of the IEEE conference on computer vision and
  pattern recognition}, 2017, pp. 1765--1773.

\bibitem{baluja2017adversarial}
Shumeet Baluja and Ian Fischer,
\newblock ``Adversarial transformation networks: Learning to generate
  adversarial examples,''
\newblock {\em arXiv preprint arXiv:1703.09387}, 2017.

\bibitem{xiao2018generating}
Chaowei Xiao, Bo~Li, Jun-Yan Zhu, Warren He, Mingyan Liu, and Dawn Song,
\newblock ``Generating adversarial examples with adversarial networks,''
\newblock {\em arXiv preprint arXiv:1801.02610}, 2018.

\bibitem{quiring2018forgotten}
Erwin Quiring, Daniel Arp, and Konrad Rieck,
\newblock ``Forgotten siblings: Unifying attacks on machine learning and
  digital watermarking,''
\newblock in {\em IEEE European Symposium on Security and Privacy}, 2018, pp.
  488--502.

\bibitem{schottle2018detecting}
Pascal Sch{\"o}ttle, Alexander Schl{\"o}gl, Cecilia Pasquini, and Rainer
  B{\"o}hme,
\newblock ``Detecting adversarial examples-a lesson from multimedia security,''
\newblock in {\em European Signal Processing Conference}. IEEE, 2018, pp.
  947--951.

\bibitem{zhang2021universal}
Chaoning Zhang, Philipp Benz, Adil Karjauv, and In~So Kweon,
\newblock ``Universal adversarial perturbations through the lens of deep
  steganography: Towards a fourier perspective,''
\newblock {\em arXiv preprint arXiv:2102.06479}, 2021.

\bibitem{zhang2016reversible}
Weiming Zhang, Hui Wang, Dongdong Hou, and Nenghai Yu,
\newblock ``Reversible data hiding in encrypted images by reversible image
  transformation,''
\newblock {\em IEEE Transactions on multimedia}, vol. 18, no. 8, pp.
  1469--1479, 2016.

\bibitem{liu2018reversible}
Jiayang Liu, Dongdong Hou, Weiming Zhang, and Nenghai Yu,
\newblock ``Reversible adversarial examples,''
\newblock {\em arXiv preprint arXiv:1811.00189}, 2018.

\bibitem{yin2019reversible}
Zhaoxia Yin, Hua Wang, Li~Chen, Jie Wang, and Weiming Zhang,
\newblock ``Reversible adversarial attack based on reversible image
  transformation,''
\newblock {\em arXiv preprint arXiv:1911.02360}, 2019.

\bibitem{yin2021reversible}
Zhaoxia Yin, Li~Chen, and Shaowei Zhu,
\newblock ``Reversible adversarial examples against local visual
  perturbation,''
\newblock {\em arXiv preprint arXiv:2110.02700}, 2021.

\bibitem{zhu2018hidden}
Jiren Zhu, Russell Kaplan, Justin Johnson, and Li~Fei-Fei,
\newblock ``Hidden: Hiding data with deep networks,''
\newblock in {\em European Conference on Computer Vision}, 2018, pp. 657--672.

\end{thebibliography}

\end{document}